%% file: main.tex
\documentclass{article}
\usepackage{spconf,amsmath,graphicx}

\input{math_commands.tex}

\usepackage{hyperref}
\usepackage{url}
\usepackage{graphicx}
\usepackage{tabularx}
\usepackage{caption} 
\usepackage{subcaption}
\usepackage{booktabs} 
\usepackage{dirtytalk} 
\usepackage{wrapfig}
\usepackage[dvipsnames]{xcolor}
\usepackage{cite} 
\newcommand{\citet}[1]{\cite{#1}}
\newcommand{\citep}[1]{\cite{#1}}

\usepackage{xparse} 
\NewDocumentCommand\squid{}{
    \includegraphics[scale=0.15]{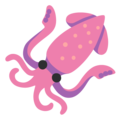}
}

\newcommand{\arkcomment}[3]{{\textcolor{#3}{[#1 #2]}}}
\renewcommand{\arkcomment}[3]{} 
\newcommand{\tbo}[1]{\arkcomment{TS:}{#1}{purple}}

\newcommand{\squiddata}[1]{SDS}

\ninept

\title{SQuId\squid: Measuring Speech Naturalness in Many Languages}

\name{
Thibault Sellam,
Ankur Bapna,
Joshua Camp,
Diana Mackinnon,
Ankur P. Parikh,
Jason Riesa
}
\address{Google
}

\begin{document}

\maketitle

\begin{abstract}
Much of text-to-speech research relies on human evaluation. This incurs heavy costs and slows down the development process, especially in heavily multilingual applications where recruiting and polling annotators can take weeks. We introduce SQuId (Speech Quality Identification), a multilingual naturalness prediction model trained on over a million ratings and tested in 65 locales---the largest effort of this type to date. The main insight is that training one model on many locales consistently surpasses mono-locale baselines. We show that the model outperforms a competitive baseline based on w2v-BERT and VoiceMOS by 50.0\%. We then demonstrate the effectiveness of cross-locale transfer during fine-tuning and highlight its effect on zero-shot locales, for which there is no fine-tuning data. We highlight the role of non-linguistic effects such as sound artifacts in cross-locale transfer. Finally, we present the effect of model size and pre-training diversity with ablation experiments.
\end{abstract}

\section{Introduction}

Evaluation is a major bottleneck for speech synthesis tasks like text-to-speech (TTS) and speech-to-speech translation. In principle, there are infinitely many spoken renditions of a piece of text, and there is no universally agreed upon definition of what makes an utterance ``correct''. Thus, researchers rely heavily on human evaluation, more specifically listening tests, during their day-to-day development cycles. The most popular type of listening test is MOS (Mean Opinion Score), during which several annotators listen to audio segments and rate them on a Likert scale between 1 to 5 (examples of foundational studies that use it include Tacotron~\citet{wang2017tacotron}, Parallel WaveNet~\citet{oord2018parallel}, or FastSpeech 2~\citet{ren2020fastspeech}). Listening tests can produce reliable results \citep{lewis2001psychometric}, since humans usually excel at detecting speech quality, and the scheme can be adapted to the need of every specific task. But they are also impractical and expensive: recruiting and polling annotators increases the cost of running experiments, slows down model research, and makes it impossible to compare results across time and institutions. The problem gets exacerbated in the multilingual setup: it may be challenging for researchers to find speakers of languages that are neither spoken by many, nor geographically close to them. Ultimately, this hinders their progress, skews the literature towards high resource languages, and prevents them from engaging in heavily multilingual research. This comes in contrast to text-based generation tasks, such as Machine Translation, for which the research community has long adopted automatic metrics (such as BLEU~\cite{bleupaper} or more recently COMET~\cite{rei-etal-2020-comet} and BLEURT~\citet{sellam-etal-2020-bleurt}) as a complement to human assessment.

To address these issues, there has been a growing interest in developing automatic metrics for speech synthesis, spearheaded by systems such as AutoMOS~\citep{45744}, MOSNet~\citep{lo19_interspeech}, or LDNet~\citep{huang2022ldnet}. The idea is to cast quality evaluation as a regression or classification problem: these systems predict a quality score from an utterance, using past listening tests as a source of training data.  The task is difficult because the target domain is complex, even in a monolingual setup: synthesis artifacts come in many forms and can affect all levels of speech production, including pronunciation, prosody, voice, and audio quality. And the task is getting harder over time~\cite{lemaguer22_interspeech}: as systems progress, the focus has shifted from obvious artifacts (e.g., robotic voices) to more subtle errors, such as inappropriate prosody or mispronunciations. Yet, the same problem that motivates the task plagues its solution---data is expensive to collect, especially outside high-resource languages, and so existing studies tend to use relatively limited training and testing sets. Early MOS Predictors have been shown to be brittle when used out of domain~\citep{cooper22Generalization}, and there are few studies outside English and Chinese.

In this paper, we study MOS Prediction at scale, and in a massively multilingual setup. We introduce SQuId (Speech Quality Identifier) a speech quality detector based on mSLAM (multilingual Speech and LAnguage Model), a recently published pre-trained model~\cite{bapna2022mslam}. SQuId is trained on over a million ratings, an order of magnitude more than most recent studies~\cite{lo19_interspeech}. More importantly it is, to the best of our knowledge, the first massively multilingual model for MOS prediction: we trained the model on 42 locales\footnote{Compared to language, locales take regional variation into account. For instance English is covered by five locales in our dataset: US English, UK English, Indian English, Nigerian English, and Australian English. Each of these variants should be rendered differently by a TTS engine.} and tested it on 65. For comparison, VoiceMOS, the most comprehensive benchmark to date, covers two locales only~\cite{huang2022voicemos}. We describe our dataset, our model, and show that SQuId outperforms a competitive baseline based on SSL and VoiceMOS by up to 50.0\%. Most improvements come from the additional supervised multilingual data, complemented by minor optimizations that target the multilingual case.  We then conduct several studies to highlight the factors that contribute to MOS prediction quality in this massively multilingual, \emph{in the wild}, setup. Key insights from our work include:

(i) Training \emph{one} model on a diverse dataset consisting of data from \emph{many} locales consistently outperforms the monolingual approach as a result of \emph{cross-locale transfer}, an effect well known in NLP~\cite{zoph2016transfer,aharoni-etal-2019-massively, Arivazhagan2019MassivelyMN, conneau-etal-2020-unsupervised}, ASR~\cite{lin2009study,li2018multi,kannan2019large} and TTS literature~\cite{Zhang2019LearningTS,he2021multilingual}. The most spectacular manifestation of this phenomenon is the model's strong performance on \emph{zero-shot} locales, where there is no labelled MOS prediction data. Cross-locale transfer allows us to increase the model's language coverage dramatically and has significant implications on evaluation of multilingual speech synthesis.

(ii) We conduct analyses to understand the nature and mechanism of cross-locale transfer for MOS prediction. We demonstrate that locale diversity has a large influence on model's performance during fine-tuning, but transfer is driven less by language similarity and more by the presence of language-agnostic phenomena (possibly including diversity of audio quality, voices and TTS systems) in the dataset. We highlight the role of \emph{para-lingual transfer}, by which training data in one locale improves performance in another for reasons orthogonal to linguistics.

(iii) Through ablation studies, we highlight the importance of modeling choices, including \emph{multilingual} pre-training and model capacity on SQuId's performance.

\textbf{Additional Related Work.}  Automatic MOS prediction has a long history in the TTS literature~\cite{ragano2022comparison}. In addition to the systems cited above, recent work includes~\cite{williams2020, mbnet21,tseng2021utilizing,choi2021neural,tian22d_interspeech}, none of which tackle multilinguality. Our work directly builds upon \citet{cooper22Generalization}, which experiments with wide range of pre-trained models. Our methodology is close, but we scale up the data (from about 30K to 1,3M samples), number of locales (from 2 to 65), model size (from 300M to 600M parameters), and contribute a novel analysis of multilinguality. Cross-lingual transfer and massively multilingual NLP have a rich history in MT~\cite{zoph2016transfer, aharoni-etal-2019-massively, Arivazhagan2019MassivelyMN} and pre-trained models~\cite{conneau-etal-2020-unsupervised, xue-etal-2021-mt5}. Authors have studied cross-lingual transfer for at least a decade in speech recognition~\cite{swietojanski2012unsupervised,huang2013cross, li2020universal, zelasko2020sounds, feng2021phonotactics} and TTS~\cite{javkin1989multilingual, Zhang2019LearningTS, Sanchez2022}.

\section{Multilingual MOS Prediction in the Wild}
\label{sec:data}


We wish to predict \emph{MOS Naturalness} ratings for both human and synthetic speech. Broadly speaking, MOS Naturalness describes how human-like an utterance sounds. Our main resource is an in-house corpus, aggregating approximately 1.9 million ratings in 66 locales across 2,092 research and commercial annotation projects completed between January 2021 and March 2022.
Most of the audio is generated  by TTS systems including both concatenative and neural systems, and the annotators are asked to select a rating based on how natural or unnatural the samples sound. The test sets used for the evaluations are primarily focused on TTS applications such as virtual assistant responses, driving directions, book passages, and news articles; general text from web-crawled corpora is also used. Sentences are typically rated in isolation (that is, outside of the context in which they originally appeared), though entire paragraphs are occasionally rated as well.  Listening tests were conducted using crowd-sourced raters on an internal ratings platform, using a 9-Point Likert Scale using 0.5 points increments This variety in test sets and TTS technologies means the stimuli contain a diverse set of errors, including pronunciation errors, text normalization errors, unnatural prosody, and acoustic artifacts such as discontinuities and signal-correlated noise.



We apply two splits to the dataset. First, we split the data by time. We use the ratings collected between January 1st and December 1st 2021 for training and development, and the rest for test. The motivation is to simulate a realistic use case, whereby a TTS engineer would want to use the past annotations to predict future ones. The second split is based on region: we hold out 24 locales for which we have exceptionally few ratings (less than 8,000 each, adding up to about 5\% of the data) and use them for test. The rationale is that small tasks yield little improvements during training but are particularly useful for analysis. Since there is no training data, we refer to these locales as \emph{zero-shot locales}, as opposed to \emph{fine-tuned locales}. The dataset is skewed towards US English (18\%), followed by UK English (12\%), and ES Spanish (4.2\%). To build the development set, we sample 2.5\% of the training set without replacement. Table~\ref{tab:datastats} provides additional statistics.


\noindent  \textbf{Challenges and Caveats.} Due to the nature of splits, we do not expect the data to be i.i.d.---the TTS systems and annotators used to produce and rate the utterances in February 2021 are usually not those of January 2022. The of number, listening conditions and quality of the annotations also vary across projects and locales, as do the input texts chosen to test the systems. Furthermore, it is generally understood that the term \textit{naturalness} is underspecified and may be interpreted differently be different raters \citep{dall2014rating,wester15c_interspeech}. 
In short, these conditions reflect TTS evaluation ``in the wild''. 
 \begin{table}[t!]
 \center
 \footnotesize
\begin{tabular} {l | c }
\toprule
Num. training utt. / systems. / locales & 969,589 / 1,476 / 42\\
Num. dev utt. / systems / locales & 34,042 / 1,474 / 42\\
Num. test utt. / systems / locales & 381,323 / 605 / 65\\
Ave. utterance duration (seconds) / num. ratings & 4.5s / 1.4\\
\bottomrule
\end{tabular}
\vspace*{-3mm}
\caption{MOS Prediction dataset statistics.}
\vspace*{-5mm}
\label{tab:datastats}
\end{table}
 
\section{The SQuId Model}

The most important design decision behind our study is to fine-tune a single model on all locales rather than keeping separate models. This offers convenience, since we have one model to maintain rather than 65. More importantly, we assume that if the model has enough capacity, \emph{positive transfer} will emerge between the locales~\cite{Arivazhagan2019MassivelyMN}.

SQuId is based on mSLAM~\citep{bapna2022mslam}, a recently published multi-modal pre-trained model trained on unlabelled speech (429K hours in 51 languages), text (15TB in 101 languages), and speech-text pairs (2.3K hours). We chose this model because it produced state-of-the-art results in many languages at the time of writing. It is based on the Conformer architecture, with 600M parameters by default. SQuId's input is a 16KHz utterance's spectrogram, along with an optional locale tag. The output is a scalar. We fine-tune the model end-to-end with a simple regression loss. After optional resampling to 16KHz, we compute an 80-dimensional log Mel spectrogram and extract mSLAM embeddings $\mathbf{e}_1^i, ,...,\mathbf{e}_T^i$ for each time step.
We mean-pool across the time dimension, returning an embedding $\mathbf{e}_{*}^i$, and apply a fully connected layer to obtain the prediction $\hat{y}^i = \mathbf{M}.\mathbf{e}_{*}^i + \mathbf{b}$. By default we use $T=3,200$ time steps, and the embeddings have dimension 1,024. The target MOS ratings are linearly rescaled from [1, 5] to [0, 1]. By default we use batch size 32 and learning rate $10^{-5}$, obtained with hyper-parameter search during a preliminary set of experiments. We train the models for 100k steps, save a snapshot every 10k steps, and export the version that yields the best version on our development set. We run experiments on Cloud TPU v3, using the Adam optimizer~\tbo{to fact-check} with 1,500 warmup steps.

Additionally, two optimizations lead to slight but consistent performance improvement. (i) We embed the locale tags of each utterance into a 64-wide vector $\mathbf{e_\ell}^i$ and concatenate it to $\mathbf{e}_{*}^i$, forming vector $[\mathbf{e}_{*}^i, \mathbf{e_\ell}^i]$. For 5\% of the data, we use a wildcard identifier \textsc{Any-Loc}, which we use for inference on locales unseen during training. (ii) We sample the data with temperature to rebalance the relative proportion of the training locales. As described in~\cite{Arivazhagan2019MassivelyMN}, we resample each locale $\ell$ with probability $p_\ell^{1/\tau}$. We use $\tau=10$, obtained by hyper-parameter search on the development set.

\section{Performance}

Let us now present SQuId's overall performance. To validate our approach, we first ensure that it performs well on VoiceMOS'22, currently the main benchmark for MOS prediction. We then scale up the test set and analyze its performance on 65 locales. Throughout the section, we will compare SQuId to Big-SSL-MOS, a competitive baseline in the spirit of SSL-MOS~\citet{cooper22Generalization}. We fine-tune w2v-BERT on the main VoiceMOS dataset with a regression objective, using a 600M parameters version of the pre-trained model to ensure a fair comparison (\citet{cooper22Generalization} uses up to 317M). The model comes in two variants: English-only, and multilingual. The architecture and dataset of our w2v-BERT implementation are described in detail in \citet{bapna2022mslam}.

For all experiments we report the correlation with human ratings. We use the segment-level Kendall Tau---segment-level because there are many locales for which we only have a handful of systems so reporting system-level correlations would be too brittle, and Kendall Tau because it is resistant to outliers (it is based on rank) and has a rich history in the meta-evaluation literature~\citep{huang2022voicemos, ma-etal-2019-results}. By default, all experiments are replicated three times and the results averaged, confidence bars represent bootstrapping-based 95\% intervals.

\begin{table}[!t]
\centering
\footnotesize
\begin{tabular} {l  l | c }
\toprule
Model & & Kendall Tau\\
\midrule
VoiceMOS '22& SSL-MOS & 0.690\\
Baselines& MOSA-Net & 0.621\\
& LDNet & 0.599\\
Submissions& Top & \textbf{0.730} \\
& Median & 0.698 \\
& Last & \emph{0.562} \\
\midrule
Big-SSL-MOS & English & \textbf{0.702}\\
 & Multilingual & 0.693\\
\midrule
SQuId & SQuIdDS & 0.606\\
& VoiceMOS & 0.700\\
& SDS+VMOS & \textbf{0.701}\\
\bottomrule
\end{tabular}
\vspace{-2mm}
\caption{Results on VoiceMOS'22, Main Track. Segment-level correlation with human ratings (Kendall Tau). The VoiceMOS'22 results were obtained from~\cite{huang2022voicemos}, except SSL-MOS which we recalculated to confirm the alignment.}
\label{table:voicemos}
\vspace{-6mm}
\end{table}

\noindent \textbf{Correctness Check: VoiceMOS'22 - Main Track.}
VoiceMOS is the main public benchmark for MOS prediction. The main track involves 4,974, 1,066, and 1,066 utterances for training, development, and test respectively, all in English. The submissions are aggressively optimized towards the task; for instance UT-MOS~\citep{UTMOS22}, a top performing system, ensembles several models and use contextual information such as Listener and Domain ID. Additionally, the organizers provide the results of three recently published baselines: SSL-MOS~\citep{cooper22Generalization}, MOSA-NET~\citep{zezario2021deep}, and LDNet~\citep{huang2022ldnet}. Since the dataset is small comparatively (we test on two orders of magnitude more data in 30X more locales), we use it as a correctness check for our approach and baselines. We restrict ourselves to simple approaches: we use batch size 8 (instead of 32) and train for 10K steps (instead of a million), but do not introduce any additional optimization. Table~\ref{table:voicemos} presents the results. We present three versions of SQuId: fine-tuned on the SQuId Dataset (SQuIdDS), VoiceMOS, and both sequentially. Although our approach is simple, we confirm that it is competitive. SQuId models trained on VoiceMOS data outpeform all three VoiceMOS baselines, and would sit in the upper half of the benchmark. Our baseline Big-SSL-MOS performs even better, justifying that the comparison is fair.

\begin{figure}[b!]
  \centering
   \vspace{-3mm}
    \includegraphics[width=0.7\linewidth]{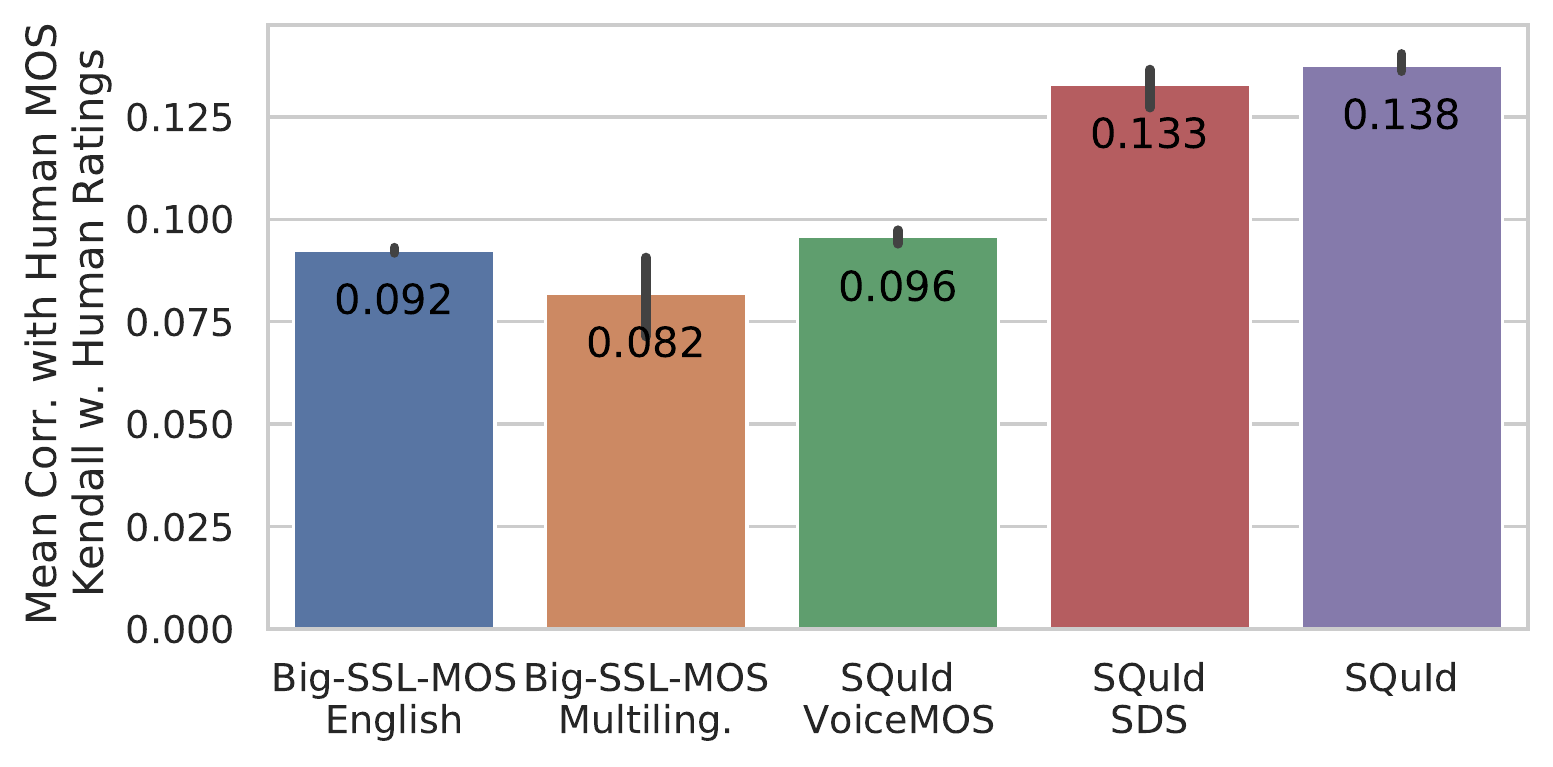}
    \vspace{-3mm}
    \caption{Correlations with human ratings, all locale averaged.}
    \label{fig:squid_scores-compact}
\end{figure}

\noindent \textbf{Main Results.} Figure~\ref{fig:squid_scores-compact} reports SQuId's performance on the main dataset. We compute the correlation with human ratings on all 65 test locales and average the results. SQuId VoiceMOS is slightly better than Big-SSL-MOS English; the main difference is due to the switch from w2v-BERT to mSLAM, which performs better on this task. Using the SQuId Dataset incurs a dramatic ~38\% performance boost, which validates the usefulness of the dataset. The optimizations (re-sampling and locale identifier) add a more modest 3.75\% increase. Our final model outperforms the best baseline by ~50.0\%, with superior results on 60 locales out of 65.

\begin{figure}[t!]
  \centering
    \includegraphics[width=0.65\linewidth]{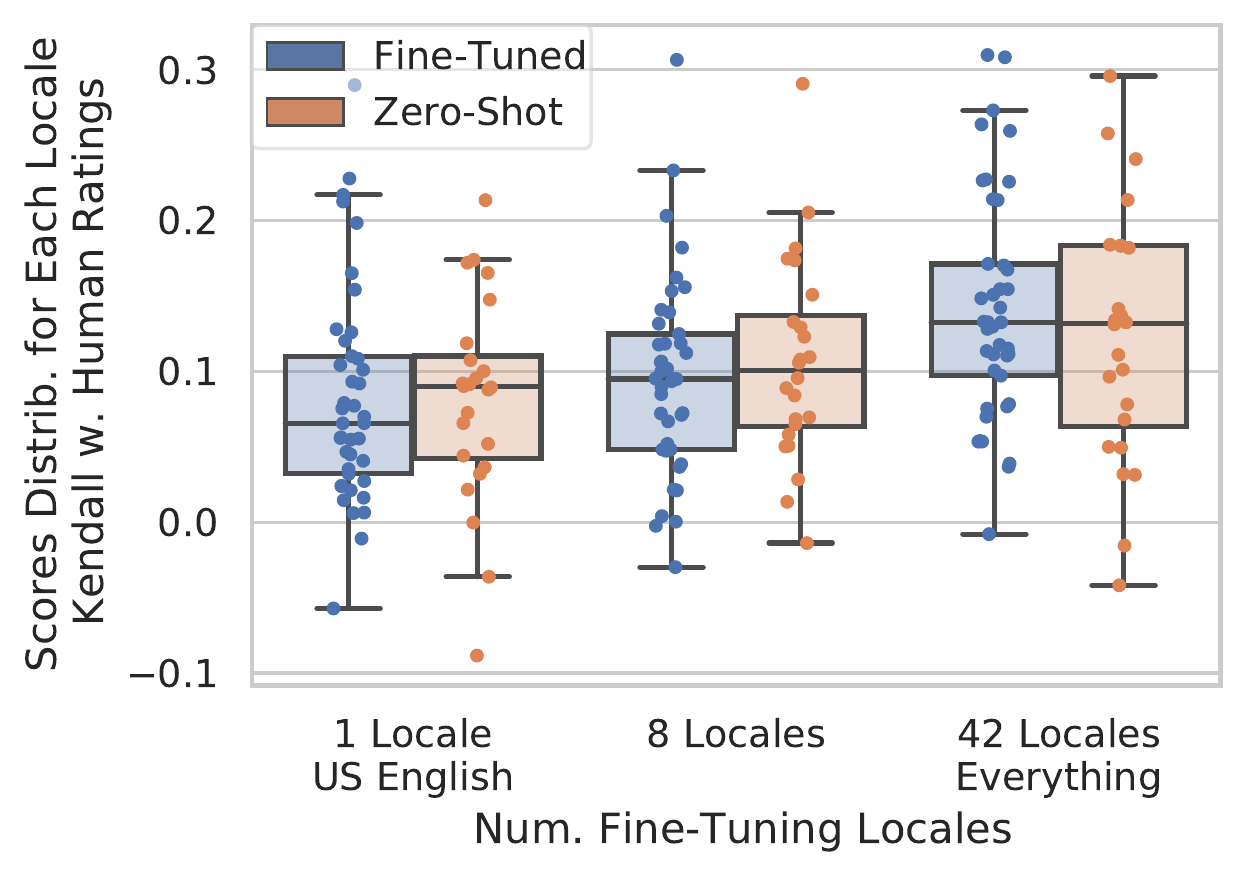}
    \vspace{-4mm}
    \caption{Distributions of the scores. Each dot represent a locale, the box plots report quartiles.}
    \vspace{-5mm}
    \label{fig:squid_scores_by_group}
\end{figure}

Figure~\ref{fig:squid_scores_by_group} presents SQuId's performance on each locale, as we increase the number of locales used for fine-tuning. Each addition improves the model, on both fine-tuned and zero-shot splits, demonstrating the existence of cross-locale transfer. SQuId's performance is comparable in average on both sets, which illustrates the viability of zero-shot inference for MOS prediction.

There are large disparities between the scores, which can range from below 0 to over 0.3. A first, and possibly main explanation, is that the test set is not homogeneous: the sub-tasks have various degrees of complexity and the annotations have various degrees of noisiness. Indirect evidence is that Big-SSL-MOS and SQuId tend to perform similarly on the same locales even though they were fine-tuned on very different datasets (Pearson correlation: 0.845). Furthermore, the training data is also heterogeneous, and so we cannot expect its performance to be constant, even on a perfect test set.

\section{Understanding Cross-Locale Transfer during Fine-Tuning}
\label{sec:abl8-finetuning}

\begin{figure}[b!]
    \centering
    \vspace{-4mm}
    \includegraphics[width=\linewidth]{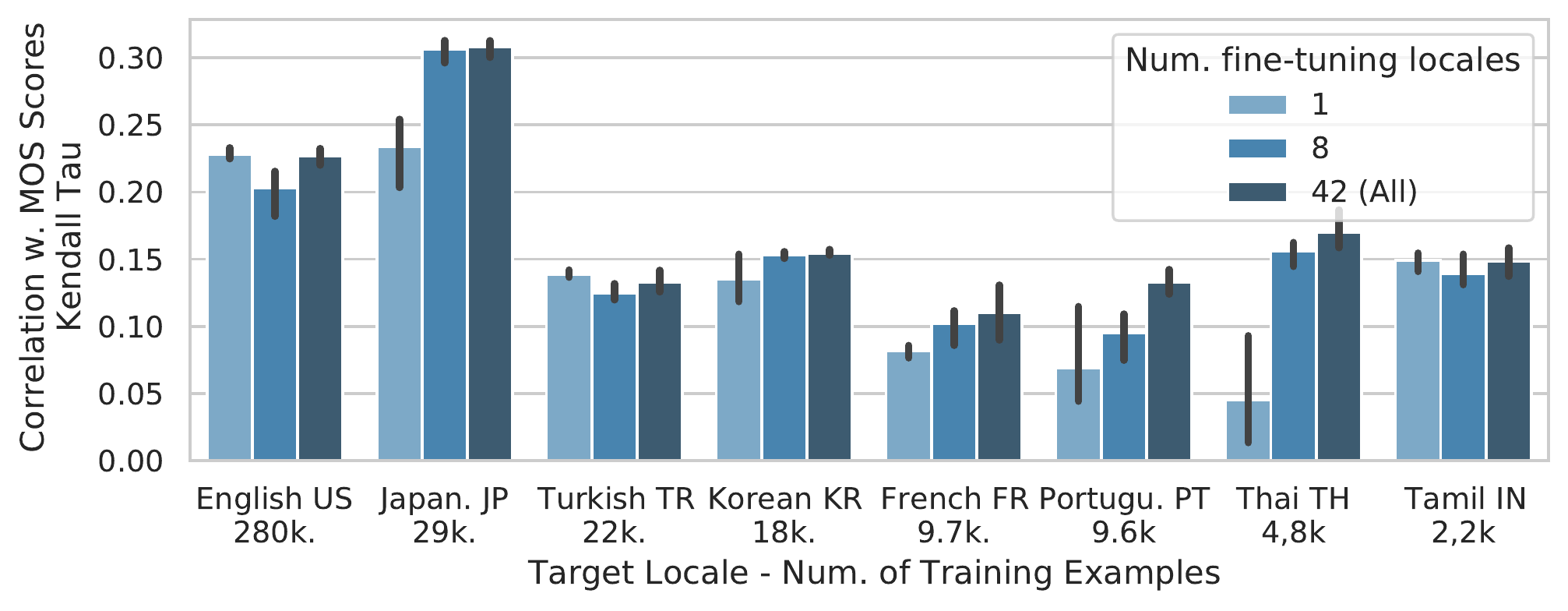}
    \vspace{-7mm}
    \caption{Performance on 8 target locales varying the number of fine-tuning languages.}
    \label{fig:vary_ft_locales}
\end{figure}
\noindent \textbf{Measuring Cross-Locale Transfer.} To understand cross-locale transfer further, we focus on 8 languages and compare three models: mSLAM fine-tuned on the target locale only, on all 8 locales, and on all the locales we have (42). We picked languages from different families, with different levels of coverage (from 2.2K to 280K ratings), and for which we have ample test data to reduce the variance of the experiment. Figure~\ref{fig:vary_ft_locales} presents the results. For five out of eight setups, each addition increases the results. The most spectacular improvement happen between 1 and 8 on Thai (+244\%) and Japanese (+31.2\%). In contrast, adding data makes no difference or causes a slight degradation for three setups. We conclude that cross-locale ``works'' but it does not affect all locales uniformly.



\begin{figure}[t!]
     \centering
     \begin{subfigure}[b]{0.49\columnwidth}
         \centering
         \includegraphics[width=\textwidth]{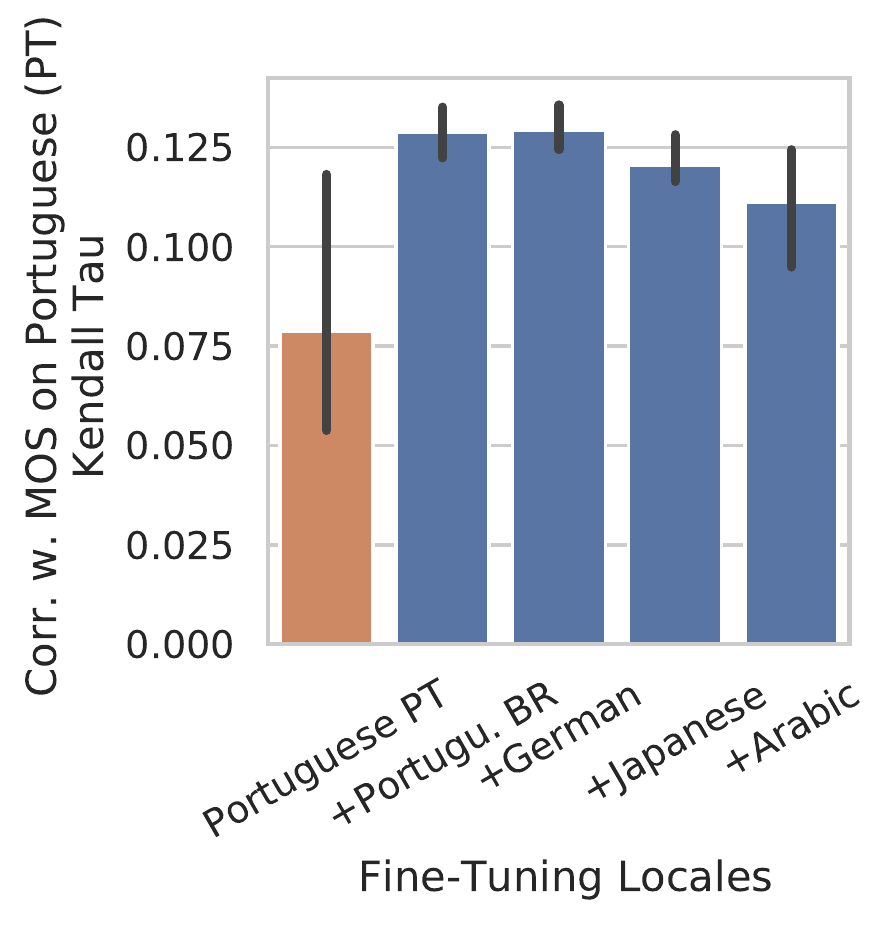}
     \end{subfigure}
     \hfill
     \begin{subfigure}[b]{0.49\columnwidth}
         \centering
         \includegraphics[width=\textwidth]{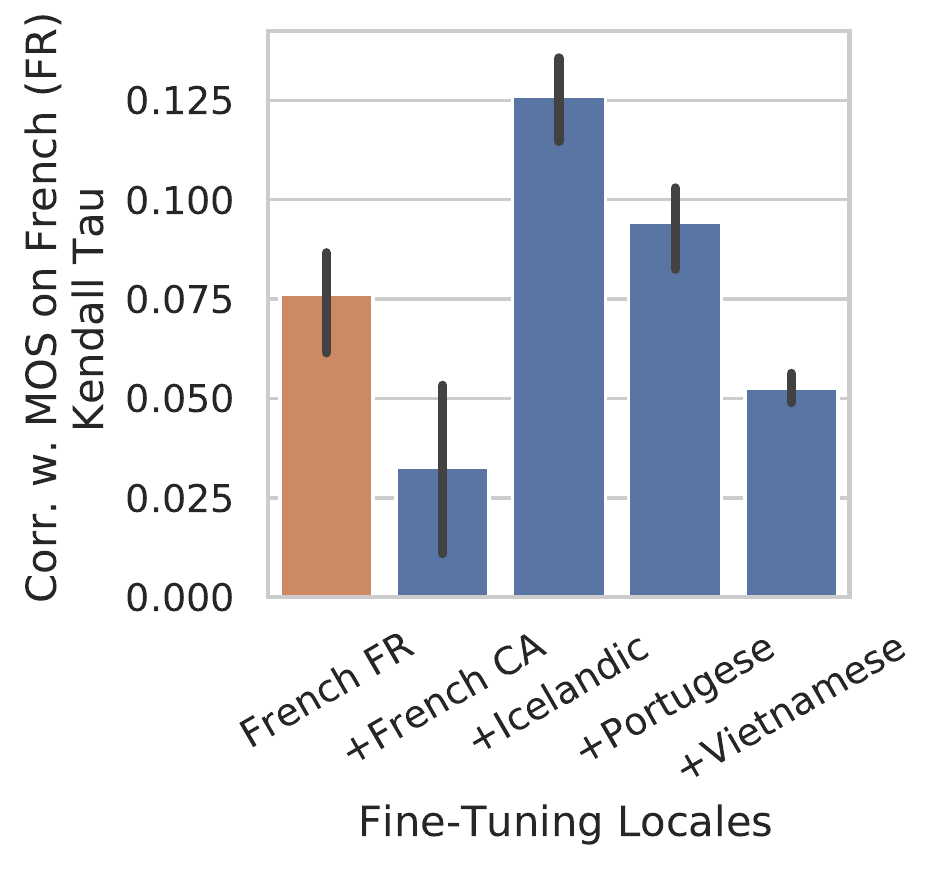}
     \end{subfigure}
     \vspace{-3mm}
     \caption{Cross-lingual transfer to Pt-PT (left) and Fr-FR (right).}
     \label{fig:langfam}
     \vspace{-6mm}
\end{figure}

\noindent \textbf{Cross-locale transfer and language.} Which locales are impacted by transfer? Intuitively, we may expect it to occur between tasks that are linguistically similar to each other~\cite{zoph2016transfer, li2018multi,kannan2019large}. To test this hypothesis, we focus on two languages for which we have different locales: Portuguese (the data set contains utterances in Brazilian and European Portuguese) and French (we have French and Canadian). We first fine-tune mSLAM on the locale with the lowest amount of data (i.e., European Portuguese and French, with about 9,700 examples each) and test it on the same. We then compare to mSLAM fine-tuned on locales pairs, by coupling the target locale with either a close one (e.g., European Portuguese and Brazilian Portuguese) or remote ones (e.g., European Portuguese and Japanese). If the hypothesis holds, then adding in-language data will help, and similar pairs will perform better than remote ones. We chose the baselines locales such that the amount of fine-tuning data is approximately similar in all setups. Figure~\ref{fig:langfam} presents the results. There is much transfer from Brazilian to European Portuguese, but the effect is approximately similar with German. Furthermore, Japanese and Arabic are not far behind. The results are more spectacular on French: adding French Canadian to European French actually \textit{decreases} its performance, while Icelandic improves it. Our conclusions are therefore negative: it seems that the influence of linguistic similarity on cross-lingual transfer is very limited in this application. A similar observation may be found in~\cite{Zhang2019LearningTS}, where improvements are observed in TTS systems transferring voices across highly dissimilar languages like English and Chinese.

\noindent \textbf{Cross-Lingual vs. Para-Lingual Transfer.}  If language similarity does not facilitate transfer, then what does? We hypothesize that the task is in fact dominated by sound and paralinguistics, i.e., artifacts such as robotic voices and flat prosody that do not depend on words. This does not mean that there is no cross-locale transfer: the performance on each task does indeed improve as the model is exposed to the others. But the phenomenon is distinct from the cross-lingual transfer that is commonly discussed in the NLP literature~\cite{zoph2016transfer}. We distinguish \emph{para-lingual} transfer from \emph{cross-lingual} transfer, and treat them as two orthogonal components of \emph{cross-locale transfer}.





\begin{figure}[t!]
    \centering
    \vspace{-3mm}
    \includegraphics[width=0.3\textwidth]{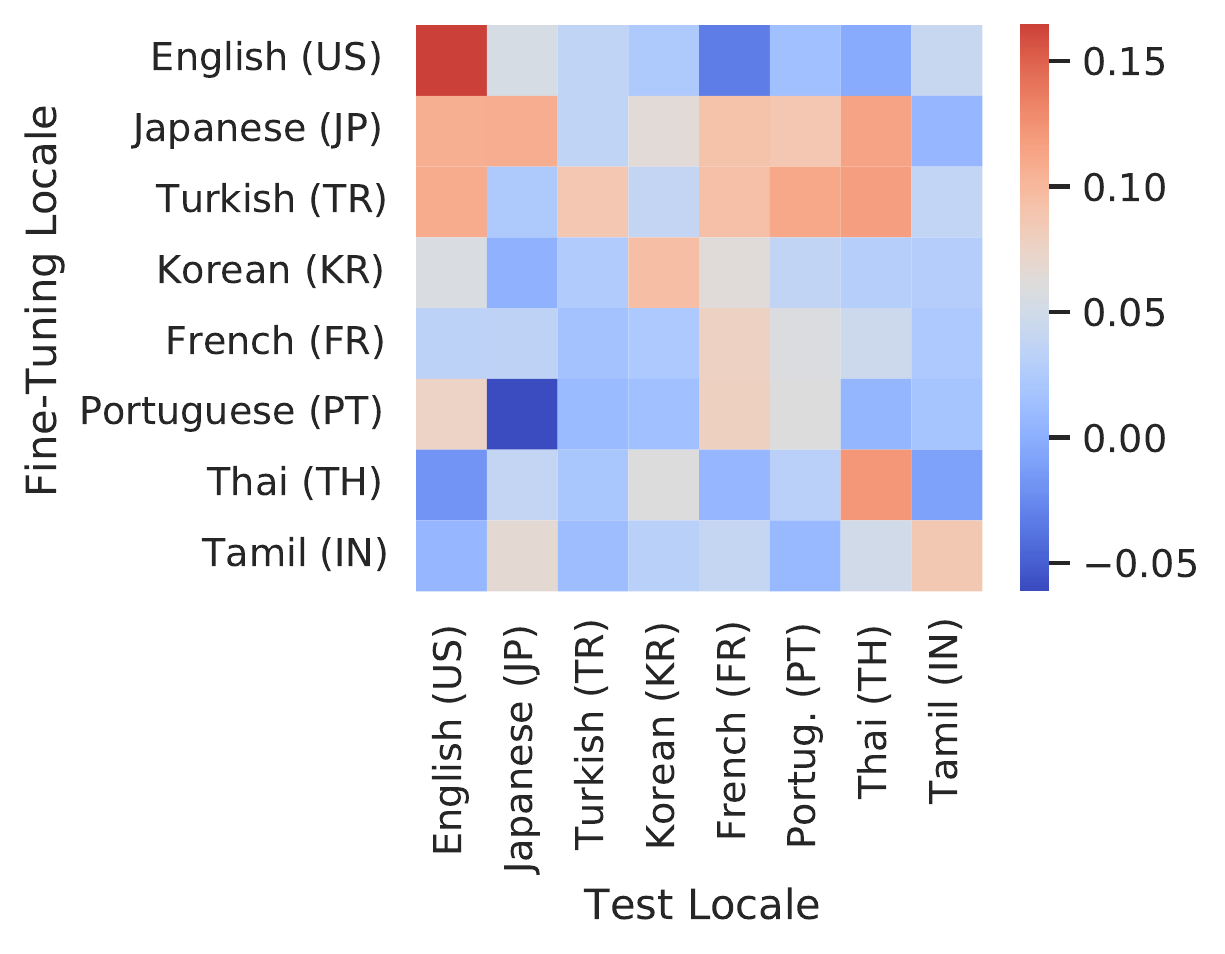}
    \vspace{-3mm}
    \caption{Cross-locale performance, Kendall Tau with human ratings.}
    \label{fig:cross-locale}
    \vspace{-5mm}
\end{figure}

To highlight the role of para-lingual transfer, we fine-tune single-locale mSLAM models on the 8 locales discussed above and run them on each other's test set, i.e., we run each of the 8 models on each of the the 8 locales. We present the results in a matrix where the rows represent fine-tuning and columns test locales (Figure~\ref{fig:cross-locale}). We trained randomly initialized models to prevent any cross-lingual transfer, with a smaller architecture to stabilize the learning (we used the 42M parameters model discussed in the next section). If the tasks were truly independent, the diagonal would be salient. If they depended on language alone, there would be more affinity between languages of the same family than between very different ones, or at least, between languages that share phonological features.
Instead we find that the results are diffuse: the good results of Turkish on Portuguese could come from a shared phonemic inventory, but the fact that Japanese yields relatively good results on five out of eight tasks (doing better than French on French) confirms that MOS Naturalness is, to some extent, orthogonal to language.

\textbf{Transfer between Pre-Training and Fine-Tuning} If the task is dominated by para-linguistics and sound artifacts, does it still make sense to pre-train models in multiple languages, and do we still benefit from scale? We rerun our experiments varying the sets of pre-training languages (from 1 to 51) and capacities (from 42M to 600M parameters), using speech-only models trained on  VoxPopuli~\cite{wang-etal-2021-voxpopuli} to simplify the setup.

\begin{figure}[h]
     \centering
     \vspace{-2mm}
     \begin{subfigure}[b]{0.49\columnwidth}
         \centering
         \includegraphics[width=\textwidth]{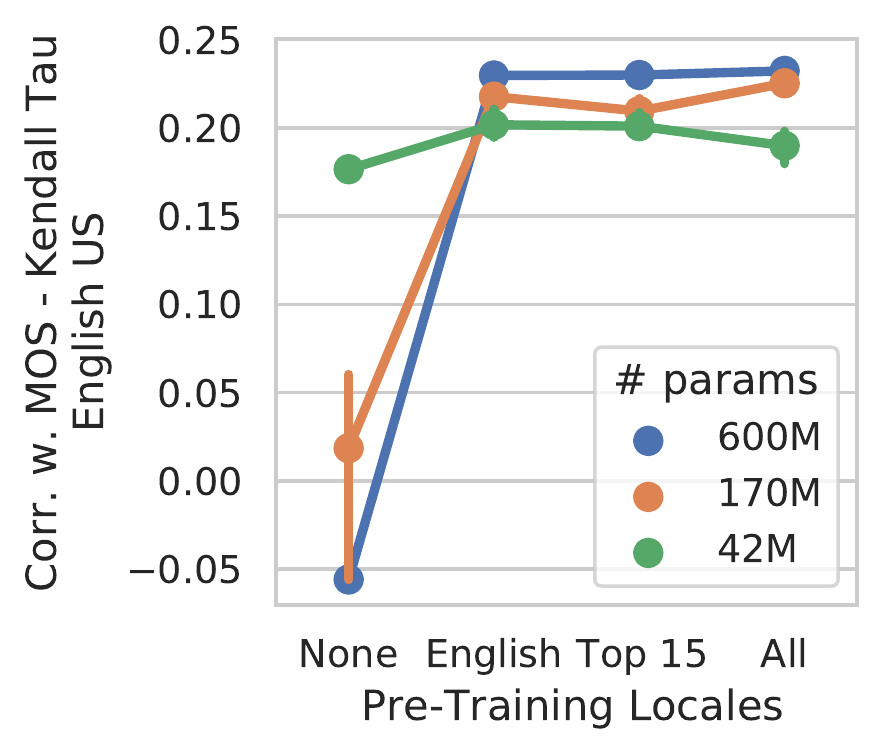}
     \end{subfigure}
     \hfill
     \begin{subfigure}[b]{0.49\columnwidth}
         \centering
         \includegraphics[width=\textwidth]{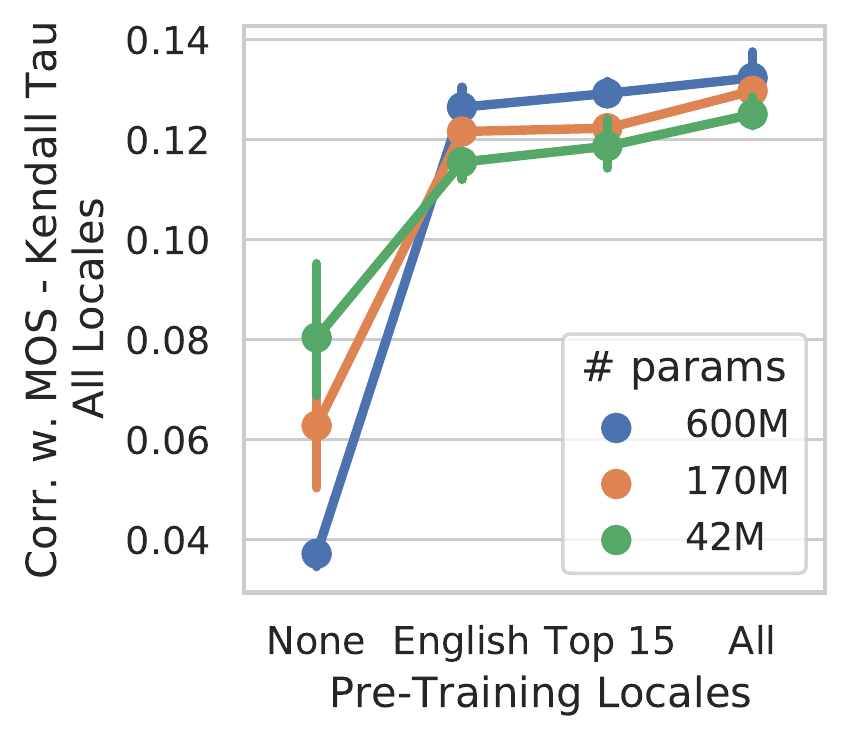}
     \end{subfigure}
     \vspace{-3mm}
     \caption{Impact of model capacity and number of pre-training locales on SQuId's performance.}
     \label{fig:pretraining}
     \vspace{-3mm}
\end{figure}

\noindent Figure~\ref{fig:pretraining} presents the results. On US English (left), pre-training in English seems essential, but any addition causes little to no improvement. The smallest model actually gets worse, we hypothesize that it is saturated (see the curse of multilinguality in the NLP literature~\cite{conneau-etal-2020-unsupervised}). The model however improves consistently if we broaden the analysis to all locales. Here again, it is difficult to assess whether the improvement comes from in-locale, cross- or para-lingual transfer, but it is clear that linguistic diversity improves performance, as does model size.



\section{Conclusion}
Zero-shot MOS prediction seems to be a viable path towards lowering the cost of massively multilingual speech synthesis evaluation. However, cross-locale transfer may in fact have little to do with language similarity. Our study holds lessons for both TTS evaluation and SSL. For TTS evaluation, we see that naturalness is only one facet of speech quality, and reiterate the importance of using complementary evaluation methods, both human and automatic, to measure language-specific aspects such as prosody and pronunciation. Regarding SSL, it seems plausible that para-locale transfer affects other tasks well, such TTS and ASR. We therefore call for model developers to enrich their pre-training corpora to include a diverse set of voice (natural and synthetic), prosody, and sound~\cite{hsu2021robust}, in addition to variations in domains and languages.~\tbo{Too weak?}

{\footnotesize
\bibliographystyle{IEEEbib}
\bibliography{main}}

\newpage
\appendix
\section{Training Data and Scores}
\label{sec:training_data_details}

\begin{figure}[ht!]
    \centering
    \includegraphics[width=0.8\columnwidth]{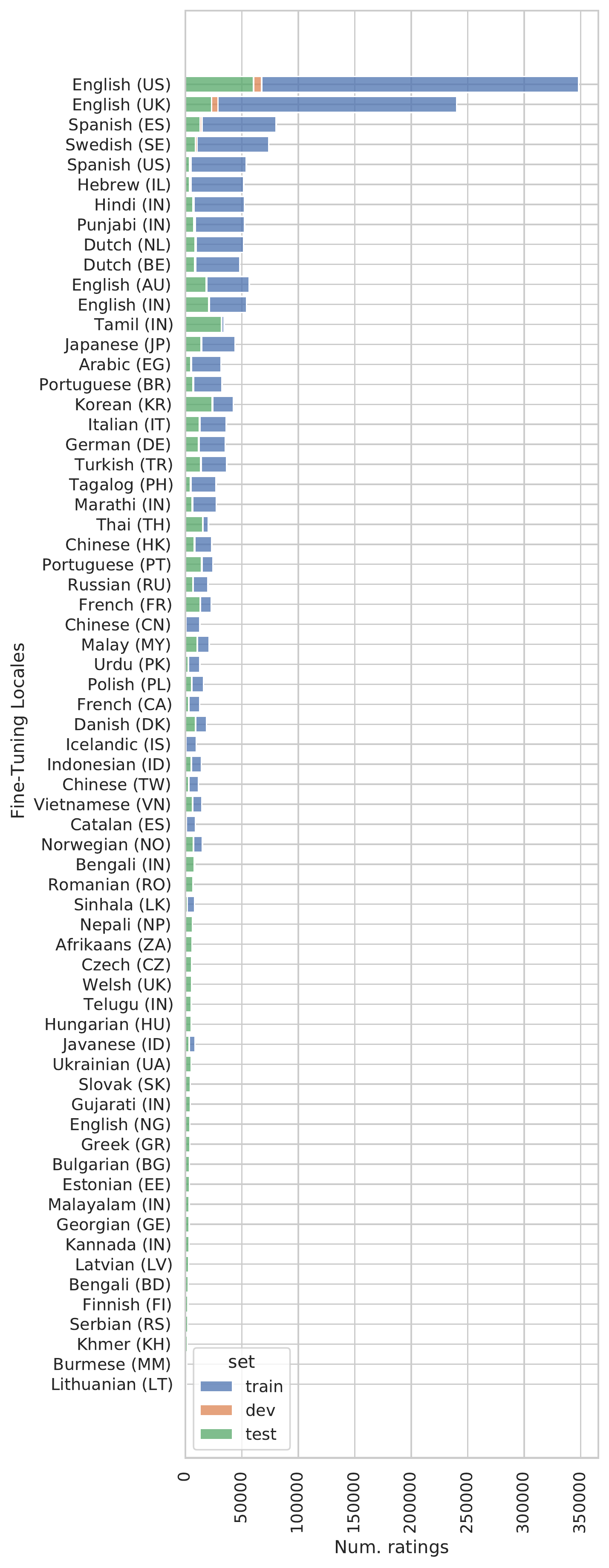}
    \caption{Distribution of data for training, development, and test in each locale.}
    \label{fig:data_distrib}
\end{figure}

\begin{figure}[ht!]
    \centering
    \vspace{20mm}
    \includegraphics[width=1.0\columnwidth]{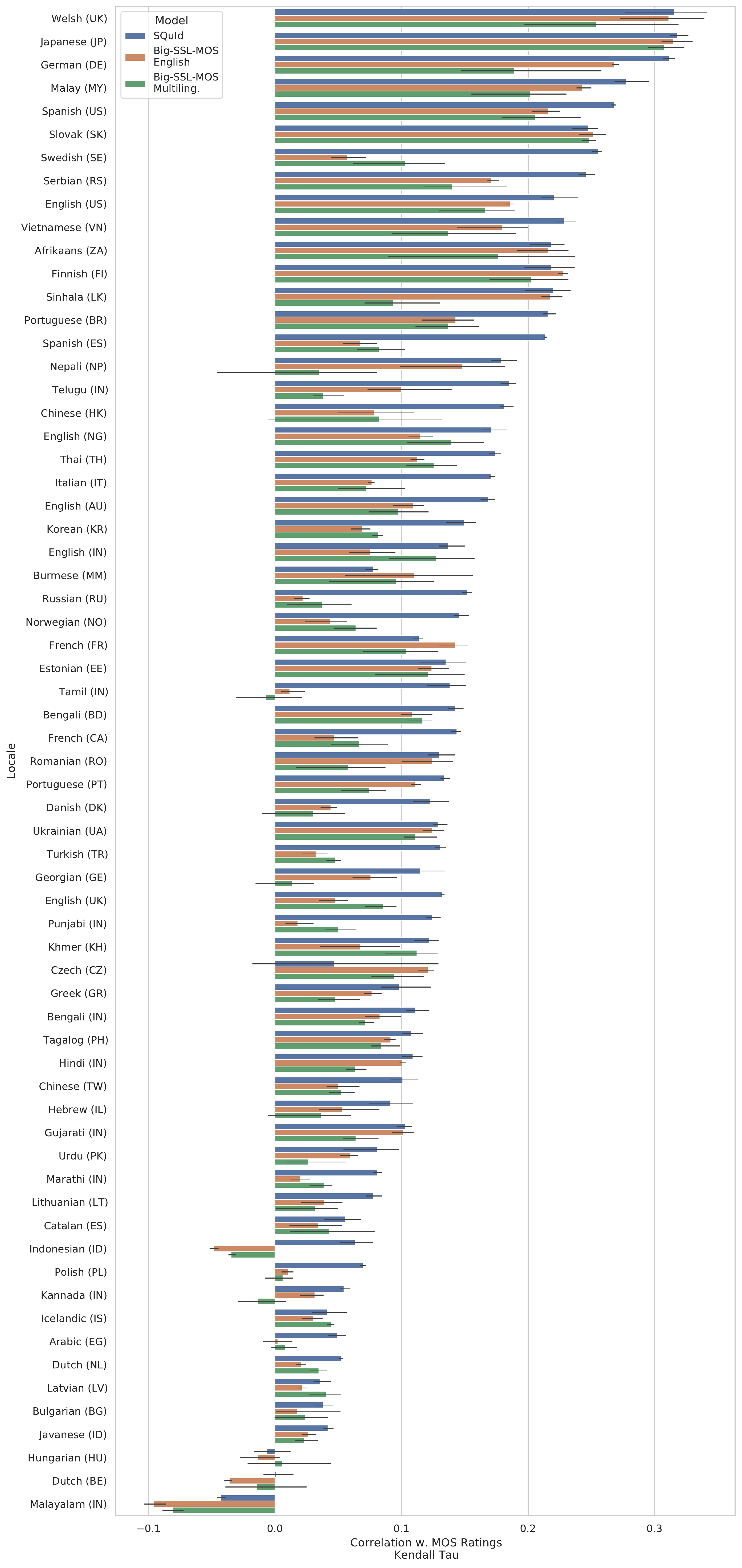}
    \caption{SQuId's performance on each locale. Each experiment was run 3 times, the confidence bars represent 95\% confidence intervals obtained with bootstrapping.}
    \label{fig:locales_scores}
    \vspace{5mm}
\end{figure}

Figure~\ref{fig:data_distrib} presents the distribution of each locale in the train, development, and test set. Figure~\ref{fig:locales_scores} presents SQuId's scores on each test locale.
\vspace{5mm}

\newpage
\begin{table}[!th]
\centering
\footnotesize
\begin{tabular} {l  l | c }
\toprule
Model & & Kendall Tau\\
\midrule
VoiceMOS '22& SSL-MOS & 0.664\\
Baselines& MOSA-Net & 0.616\\
& LDNet & 0.592\\
Submissions& Top Submission & \textbf{0.722}\\
& Median Submission & 0.664\\
& Bottom Submission & 0.305\\
\midrule
Big-SSL-MOS & En. & 0.678\\
& Multilingual & \textbf{0.711}\\
\midrule
SQuId & SDS & 0.368\\
 & VoMOS OOD & 0.657\\
 & SDS+VMOOD & \textbf{0.677}\\
\bottomrule
\end{tabular}
\caption{Results on VoiceMOS'22, OOD Track. Segment-level correlation with human ratings (Kendall Tau). The VoiceMOS'22 results were obtained from~\cite{huang2022voicemos}.}
\label{table:voicemos-ooo}
\end{table}
\section{Results on VoiceMOS, Out-of-Domain Track}
\label{voicemos:ood}

The VoiceMOS Challenge also includes an ``out-of-distribution'' track, in which the datasets are much smaller --- 136, 136, and 540 utterances for train, development, and test respectively, with an additional 540 unlabelled samples. The utterances are in Chinese. Table~\ref{table:voicemos-ooo} presents our results. The SQuId models fine-tuned on VoiceMOS OOD are competitive with the baselines, but they are out-performed by Big-SSL-MOS Multilingual.

\section{Relationship between training data and performance}
\label{sec:data_vs_perf}

\begin{figure}[b!]
     \centering
     \begin{subfigure}[b]{0.49\columnwidth}
         \centering
         \includegraphics[width=\textwidth]{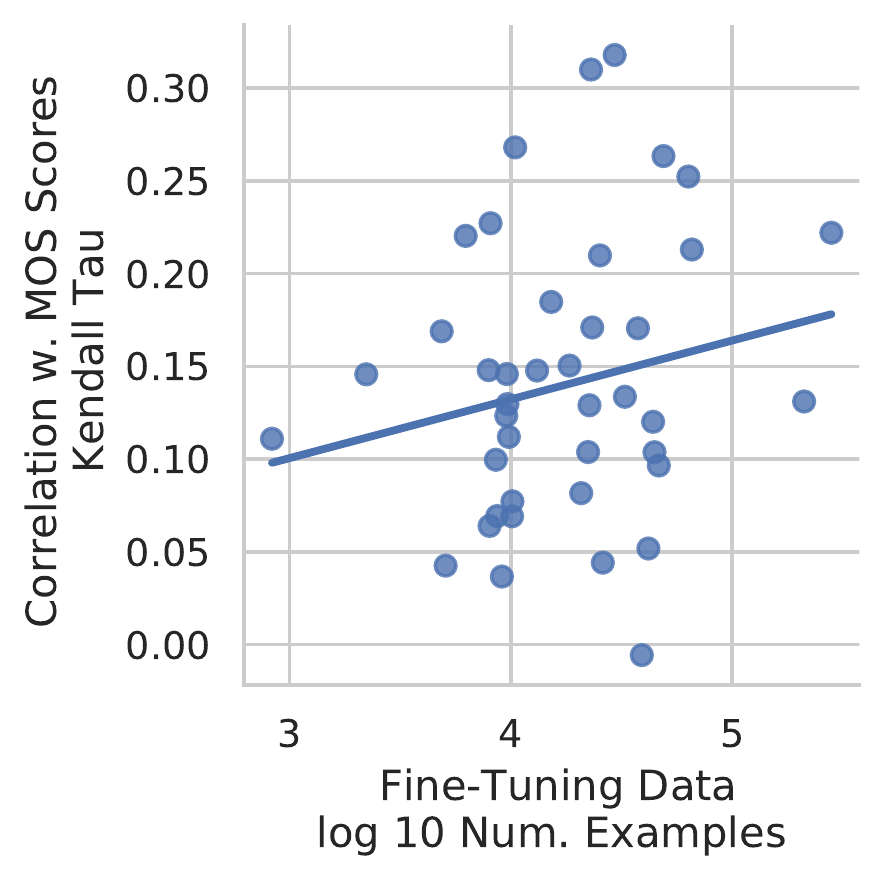}
     \end{subfigure}
     \hfill
     \begin{subfigure}[b]{0.49\columnwidth}
         \centering
         \includegraphics[width=\textwidth]{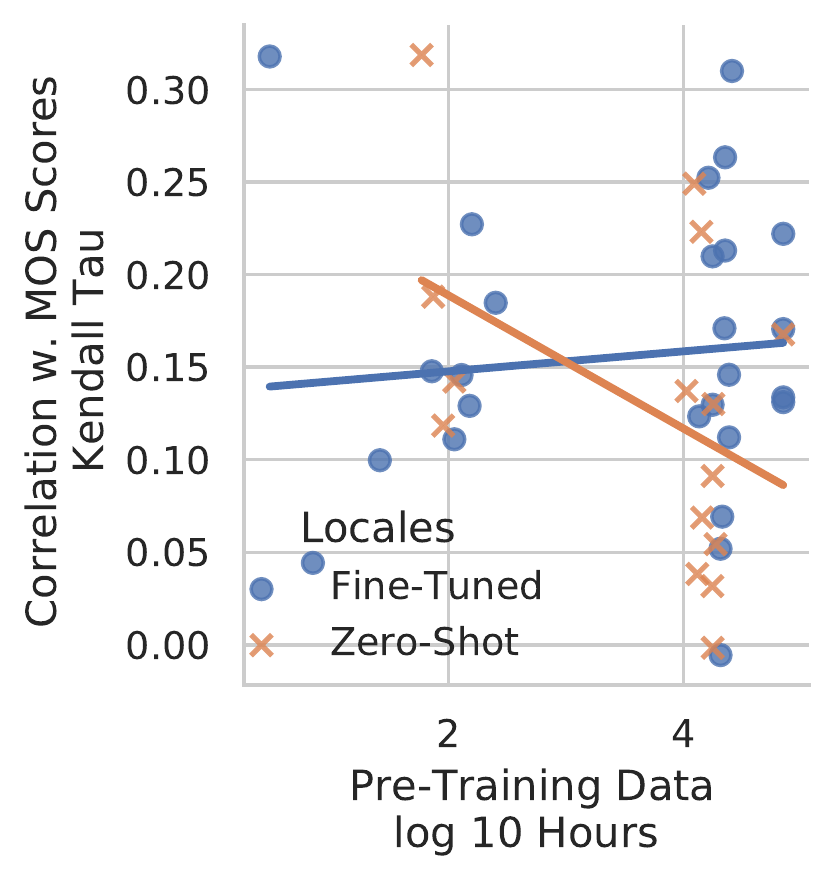}
     \end{subfigure}
     \caption{Relationship between between fine-tuning data (left) and pre-training data (right) and performance on each locale. Pearson correlation between log size and MOS scores: 0.198 for fine-tuning (left), -0.149 for pre-training (right).}
     \label{fig:data_vs_perf}
\end{figure}



What is the relationship between the amount of training data for a given locale and SQuId's performance? A naive hypothesis is that more data directly leads to performance gains, as might be observed in MT~\cite{Arivazhagan2019MassivelyMN}. Figures~\ref{fig:data_vs_perf} present the relationship for fine-tuning and pre-training separately. The effect does exist in fine-tuning, but it is weak. The correlation weak to negative in the pre-training case, even for zero-shot languages. We therefore hypothesize that for many locales, most of the model's knowledge is acquired not in-locale but through cross-locale transfer.


\begin{figure}[ht!]
    \centering
    \includegraphics[width=0.4\textwidth]{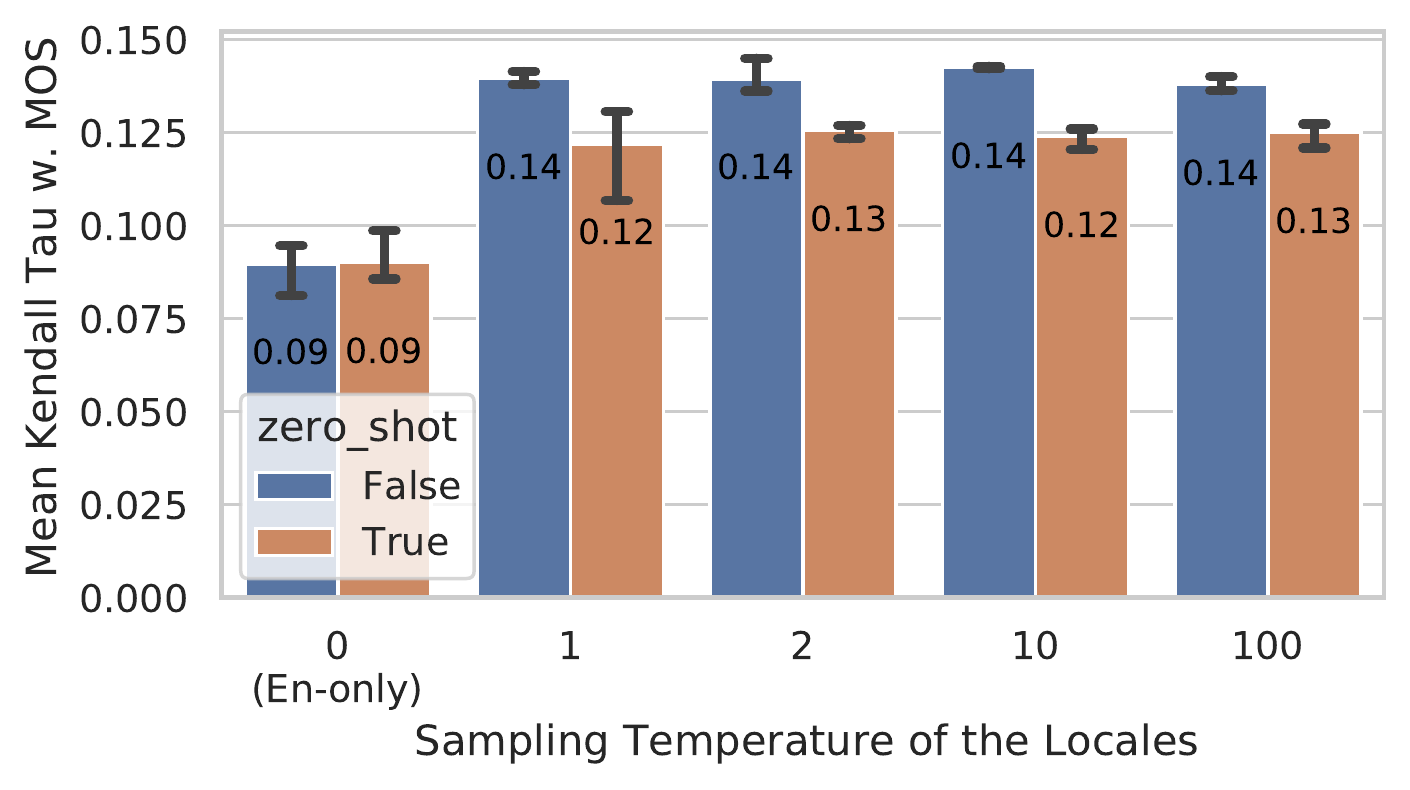}
    \caption{Impact of the sampling temperature on performance.}
    \label{fig:sampling}
\end{figure}

\section{Impact of Temperature Sampling}
\label{sec:temperature}

We study the effect of locales representation in the fine-tuning data. If $p$ the natural frequency of a locale, we resample it with probability $p^{1/\tau}$, where $\tau$ is the temperature parameter~\cite{Arivazhagan2019MassivelyMN}. A temperature $\tau=1$ means that the data is sampled according to its natural distribution (which is heavily skewed towards English), while 100 roughly corresponds to all locales being equi-probable (i.e., the high resources locales are under-sampled, while the low resource ones are over-sampled). Figure~\ref{fig:sampling} presents the results. We observe a leap when introducing the non-English languages, between $\tau=0$ to $\tau=1$. Performance on zero-shot stabilizes at $\tau=2$ (see the reduced confidence bar), then stays roughly constant, regardless of how aggressively we re-sample the data. This stability suggests that the model needs only few examples in each locale to reach its maximal performance; any repetition has little to no effect.

\section{Architectures of the Models Used in Pre-Training Ablations}
\label{sec:architectures}

Figure~\ref{fig:pretraining} presents SQuId's performance as we vary the pre-training data and number of parameters in the model. All the default hyper-parameters were obtained from the w2b-BERT model described in~\cite{bapna2022mslam}. The variations are described in table~\ref{table:hyperparameters}.

The models were trained on VoxPopuli~\cite{wang-etal-2021-voxpopuli}. The single-locale setup is based on English, the 15 locales setup is based on the 15 most represented languages, that is, English, German, French, Italian, Spanish, Polish, Dutch, Czech, Romanian, Hungarian, Greek, Bulgarian, Portuguese, Swedish, Lithuanian.

\begin{table}[!th]
\centering
\footnotesize
\begin{tabular} {l | c c c}
\toprule
Param & 42M & 170M & 600M\\
\midrule
Num. contrastive layers & 4& 4 & 8\\
Num. shared layers & 8& 8 & 16\\
Model dim. &368  & 768 & 1024\\
Num. attention heads & 4 & 8 & 8\\
Enc. kernel size & 7 & 5 & 5\\
\midrule
Batch size & 8,192 & 16,384 & 8,192\\
TPU Type & v3 & v4 & v4 \\
Training steps - 1 locale & 200K & 501K & 831K\\
Training steps - 15 loc. & 200K & 581K & 933K\\
Training steps - all loc. & 200K  & 800K & 1,300K\\
\bottomrule
\end{tabular}
\caption{Hyper-parameters of ablated mSLAM models.}
\label{table:hyperparameters}
\end{table}

\end{document}

%% file: math_commands.tex
\usepackage{amsmath,amsfonts,bm}

\def\eqref#1{equation~\ref{#1}}

\def\1{\bm{1}}

\DeclareMathAlphabet{\mathsfit}{\encodingdefault}{\sfdefault}{m}{sl}
\SetMathAlphabet{\mathsfit}{bold}{\encodingdefault}{\sfdefault}{bx}{n}

